A Deep Learning-based Method to Extract Lumen and Media-Adventitia in Intravascular Ultrasound Images

Fubao Zhu, Ph.D.[1], Zhengyuan Gao, B.S.[1], Chen Zhao, M.S.[2], Hanlei Zhu, B.S.[1], Yong Dong, MBBS[3], Jingfeng Jiang, Ph.D.[4], Neng Dai, M.D., Ph.D.[5,6], Weihua Zhou, Ph.D.[2]

[1]School of Computer and Communication Engineering, Zhengzhou University of Light Industry, Zhengzhou, Henan, China
[2]Department of Applied Computing, Michigan Technological University, Houghton, MI, USA
[3]Department of Cardiology, The 7th People's Hospital of Zhengzhou, Zhengzhou, Henan, China
[4]Department of Biomedical Engineering, Michigan Technological University, Houghton, MI, USA
[5]Department of Cardiology, Zhongshan Hospital, Fudan University, Shanghai Institute of Cardiovascular Diseases, Shanghai, China
[6]National Clinical Research Center for Interventional Medicine, Shanghai, China

*Corresponding authors:
Neng Dai, MD, PhD
Department of Cardiology, Zhongshan Hospital, Fudan University, Shanghai Institute of Cardiovascular Diseases, National Clinical Research Center for Interventional Medicine
180 Fenglin Road, Xuhui District, Shanghai 200032, China.
E-Mail: niceday1987@hotmail.com

Weihua Zhou, PhD
Department of Applied Computing, Michigan Technological University
1400 Townsend Dr, Houghton, MI 49931
E-Mail: whzhou@mtu.edu

**Abstract**: Intravascular ultrasound (IVUS) imaging allows direct visualization of the coronary vessel wall and is suitable for the assessment of atherosclerosis and the degree of stenosis. Accurate segmentation and measurements of lumen and median-adventitia (MA) from IVUS are essential for such a successful clinical evaluation. However, current segmentation relies on manual operations, which is time-consuming and user-dependent. In this paper, we aim to develop a deep learning-based method using an encoder-decoder deep architecture to automatically extract both lumen and MA border. Our method named IVUS-U-Net++ is an extension of the well-known U-Net++ model. More specifically, a feature pyramid network was added to the U-Net++ model, enabling the utilization of feature maps at different scales. As a result, the accuracy of the probability map and subsequent segmentation have been improved We collected 1746 IVUS images from 18 patients in this study. The whole dataset was split into a training dataset (1572 images) for the 10-fold cross-validation and a test dataset (174 images) for evaluating the performance of models. Our IVUS-U-Net++ segmentation model achieved a Jaccard measure (JM) of 0.9412, a Hausdorff distance (HD) of 0.0639 mm for the lumen border, and a JM of 0.9509, an HD of 0.0867 mm for the MA border, respectively. Moreover, the Pearson correlation and Bland-Altman analyses were performed to evaluate the correlations of 12 clinical parameters measured from our segmentation results and the ground truth, and automatic measurements agreed well with those from the ground truth (all Ps<0.01). In conclusion, our preliminary results demonstrate that the proposed IVUS-U-Net++ model has great promise for clinical use.

## 1. Introduction

Atherosclerosis is a disease of the vessel wall and responsible for many fatal cardiovascular diseases. Compared with the in vitro screening, the widespread application of the intravascular ultrasound (IVUS) technique relies on its capability to visualize the inner structure and the blood flow in real-time to diagnose the arteriosclerotic disease of the coronary artery. It can assess quantitative clinical measurements, such as the lumen cross-sectional area (CSA), external elastic membrane (EEM) CSA, plaque plus media CSA, and other relevant structural information of the coronary artery. However, accurate extraction of the lumen and median-adventitia(MA) border is essential for the assessment of plaque volume and stenosis degree. The current clinical practice relies on manual annotation in the IVUS frames, which is time-consuming and user-dependent.

Both traditional image processing techniques and deep learning approaches have been investigated to extract the lumen and MA border automatically. Unal et al. [1]explored the method based on the use of shape and intensity priors. Zhu et al. **Error! Reference source not found.** exploited the gradient vector flow in a nonparametric energy function to detect the target's border. Considering different characteristics of the imaging caused by radio frequency signals, Mendizabal-Ruiz et al. **Error! Reference source not found.** proposed a physics-based IVUS image reconstruction method for lumen segmentation. Recently, several techniques based on deep learning have been proposed for IVUS segmentation. Su et al. **Error! Reference source not found.** proposed a coding method that uses an artificial neural network (ANN) to classify if the pixel is located in the border. Based on the U-Net**Error! Reference source not found.**, Yang et al. designed an IVUS-Net model [6] and a Dual-Path U-Net model [7] for lumen and MA segmentation. However, the existing approaches based on U-Net all adopted a symmetrical network structure with the skip-connection between two blocks in the same deep layer. Hence, feature information at different spatial scales cannot be effectively fused and propagated over the entire network. To incorporate the multiscale feature information, in this paper, we proposed a novel method (IVUS-U-Net++) that uses the nested and dense connections**Error! Reference source not found.** to segment the lumen and MA border with the pyramid network[9]. The proposed IVUS-U-Net++ model is evaluated with human subject data from our internal imaging database.

## 2. Methodology
### 2.1. Image acquisition and description

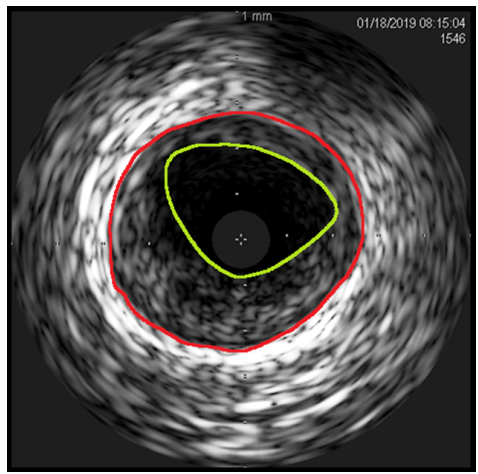

Figure 1 An example showing annotated boundaries of lumen (yellow) and MA (red) by a radiologist.

In this study, all IVUS data, including 1746 cross-sectional images from 18 patients, were collected from the 7$^{th}$ People's Hospital of Zhengzhou between February 2020 and March 2020. Each patient was scanned with a commercial IVUS machine (EagleEye, Volcano Corporation, Cordova, CA, USA) from the target lesion's distal end and pulled back to the proximal end at a fixed speed (0.5 mm per second). A well-trained IVUS analyst selected the range of slices containing the target lesion and then annotated the lumen and the MA border manually. As shown in Figure 1, the yellow and red contours represent the lumen and MA borders, respectively. Of note, in Figure 1, the region between the lumen and MA is a plaque.

### 2.2. Segmentation network

Figure 2 shows the classical symmetrical neural network architecture with the skip-connection in the classical U-Net model. In the down-sampling path, the features are captured from the images through the convolutional operators automatically. With the transpose convolution adopted in the up-sampling path, the feature maps restore the spatial structure to return to the original scale gradually. In the IVUS-Net[6], the dual-path convolution blocks, which are similar to Google's inception architecture**Error! Reference source not found.**, are employed as the convolutional blocks. This adaption combined with the multiscale information is shown in Figure 3.

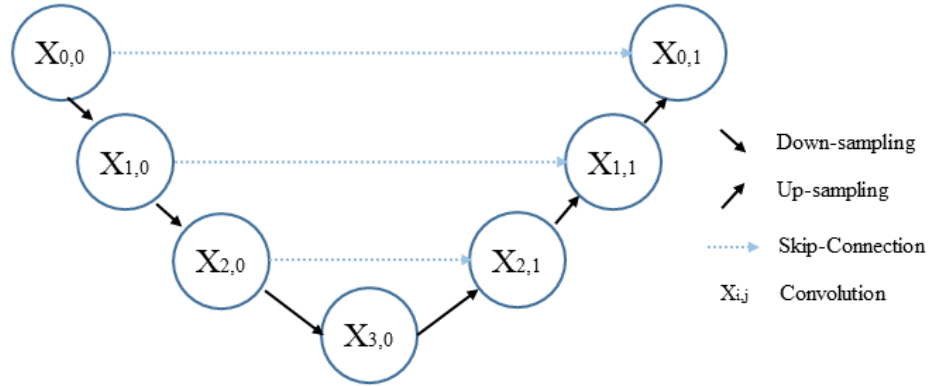

Figure 2 A schematic diagram showing a symmetrical network model with the skip-connection.

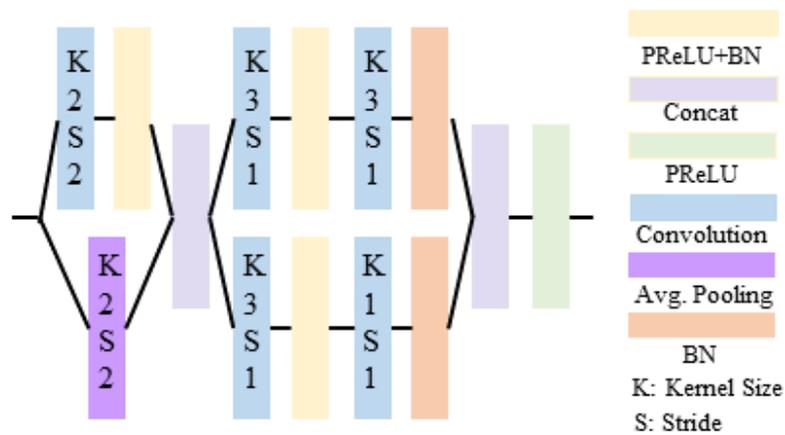

Figure 3 A diagram illustrating convolution blocks used in the IVUS-Net. This dual-path avoids information loss due to the pooling operations and produces a refined feature map to the main branch. PRelU, parametric rectified linear unit; BN, batch normalization.

In contrast, figure 4 demonstrates the architecture of our deep neural network, called IVUS-U-Net++. To better utilize the fusion of the multiscale information, a heterogeneous combination of feature maps at different spatial scales can be achieved. Of note, we use upscale operators to ensure size compatibility during the fusion of multi-scale features. Our goal is to enhance of multi-scale feature representation and fusion. For example, the feature map from the convolution block(0,3) (see Figure 4) is shown in Eq.1.

$$\mathbf{F}(i=0, j=3) = \mathbf{F}(0,0) + \mathbf{F}(0,1) + \mathbf{F}(0,2) + \mathbf{Tr}(\mathbf{F}(1,2)) \tag{1}$$

where the F(i,j) represents the feature map of the convolution block(i,j) and the Tr is to describe the convolutional transpose operator.

Several stratiges were used to accelerate the training process. First, to overcome the gradient vanishing problem from this deep network, pre-trained weights were used as the backbone**Error! Reference source not found.**. Second, batch normalization (BN)**Error! Reference source not found.** and ReLU activation function were applied after each convolution layer to accelerate network training and guarantee the non-linearity of the network. Third, to fully utilize the information extracted from different scales, five resized feature maps supervised the integrated feature map of the convolution block(0,5). By using this feature pyramid network, the final probability map was generated by a voting mechanism with this parallel connection.

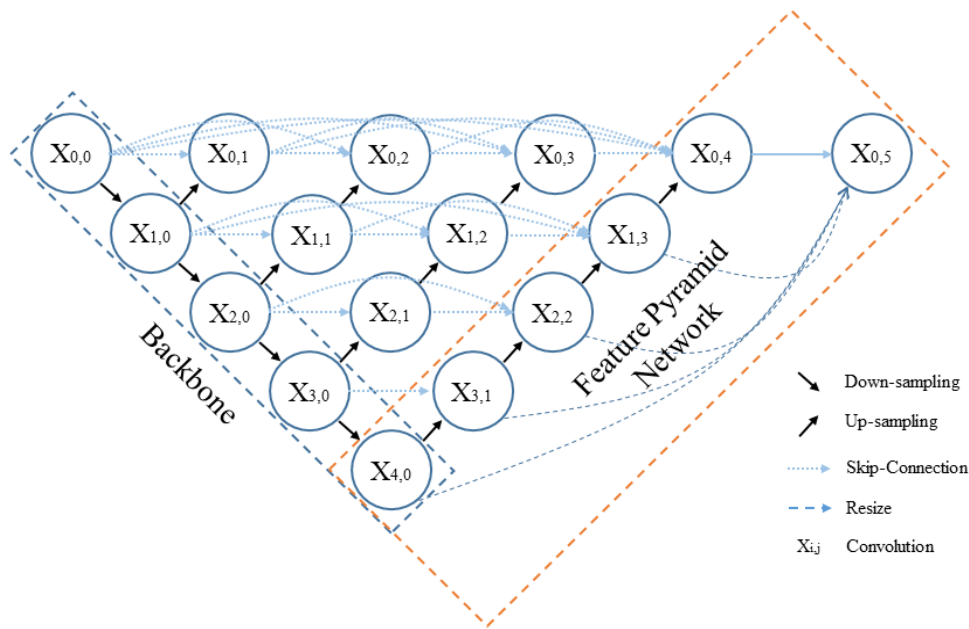

Figure 4 A plot showing network architecture of the proposed IVUS-U-Net++ model.

### 2.3. Post-processing

The proposed IVUS-U-Net++ model is trained to solve a binary classification problem. Thus, in the prediction stage, the output from the proposed IVUS-U-Net++ model is a probability map ranging from 0 to 1. Finally, the probability map is converted to the target (typically high values on the probability map) and background (typically low values on the probability map) using a self-adapting threshold, OSTU**Error! Reference source not found.**, resulting in a binary map. Using the prior such that the anatomical target only occupies one connected region, only the largest connected region from the binary map is retained and all other regions are removed**Error! Reference source not found.**.

### 2.4. Loss function

A loss function based on Dice similarity coefficient (DSC) is used as our optimization function. It penalizes the difference between the prediction output and the ground truth. The definition is shown in Eq. 2.

$$\text{DSC} - \text{Loss}(R, R') = 1 - \sum_i \sum_j \frac{2|R_{ij} \cap R'_{ij}|}{|R_{ij}| + |R_{ij}|} \times 100\% \qquad (2)$$

where $R_{ij}$ represents the ground truth in pixel (i,j). $R_{ij}$ indicates the prediction outputs of the network corresponding to the $R_{ij}$.

### 2.5 Evaluation metrics

To evaluate the effectiveness of the model, the Jaccard Measure (JM) is used to measure the consistency between the prediction binary map and the ground truth. The definition of JM is shown as:

$$\text{JM}(R, R') = \frac{|R \cap R'|}{|R \cup R'|} \times 100\% \qquad (3)$$

where the $R$ is the predicted binary map after post-processing, and the $R'$ is the ground truth.

The Hausdorff Distance (HD) is used as an evaluation metric as well. HD measures the greatest distance $(d)$ of all points belonging to the border of the predicted binary map after post-processing ($CR$) to the closest point in the corresponding border of the ground truth ($CR'$). And the $ps$ indicates the pixel spacing of the image. It is defined as in Eq. 4:

$$\text{HD}(CR, CR') = \text{MAX}\{d(CR, CR'), d(CR', CR)\} \times ps \qquad (4)$$

In addition, to further evaluate the clinical accuracy, 12 clinical parameters were calculated according to guidelines from the American College of Cardiology consensus statement on IVUS**Error! Reference source not found.**. The details of the clinical parameter are as follows.

- Maximum EEM diameter: The longest diameter through the center of the mass of the MA.
- Minimum EEM diameter: The shortest diameter through the center of the mass of the MA.
- EEM CSA: The area bounded by the MA border.
- Maximum lumen diameter: The longest diameter through the center of the mass of the lumen.
- Minimum lumen diameter: The shortest diameter through the center of the mass of the MA.
- Lumen CSA: The area bounded by the lumen border.
- Lumen eccentricity: [(maximum lumen diameter - minimum lumen diameter) / maximum lumen diameter].
- Maximum plaque plus media thickness: The largest distance from the intimal leading edge to the EEM along with any line passing through the luminal center of mass.
- Minimum plaque plus media thickness: The shortest distance from the intimal leading edge to the EEM along any line passing through the center of mass of the lumen.
- Plaque plus media CSA: (The EEM CSA - the lumen CSA.)

- Plaque plus media eccentricity: (Maximum plaque plus media thickness - minimum plaque plus media thickness) / maximum plaque plus media thickness.
- Plaque burden: Plaque plus media CSA / EEM CSA.

The longest or shortest diameter is obtained by searching all lines through the center point of mass and the area bounded by the border is calculated by counting the pixels inside the border.

## 3. Experiments and results

### 3.1. Parameter setting

We trained the models on a Linux machine equipped with a single Tesla P100 GPU with 16GB GPU memory. Our method was implemented by Keras 2.2.24 with Cuda 9.0. In addition, in the training stage, an RMSprop optimizer[16] was used to optimize the model weights with the learning rate of 0.0001 and the number of epochs was set to 201. To accelerate the convergence, the pre-trained weights in ImageNet**Error! Reference source not found.** were employed as the backbone of the encoder.

The entire dataset of 1746 images from 18 patients were randomly split into the training data of 1572 images from 18 patients for the 10-fold cross-validation and the test data containing 174 images from 18 patients for evaluating the performance of models.

3.2. Experimental results

It took 0.1203±0.0046(min 0.1152, max 0.1521) seconds for our IVUS-U-Net++ model to predict the segmentation result for each slice. To determine the optimal model and verify the robustness of the models, all the models from the 10-fold cross-validation were evaluated with the test dataset. The results can be seen in Table 1. The optimal model achieved a JM of 0.9412, a HD of 0.0639 for the lumen border and a JM of 0.9509, a HD of 0.0867 for the MA border, respectively.

Table 1 Performance of all the IVUS-U-Net++ models generated from the 10-fold cross-validation over the test dataset. The quantitative results are evaluated by Jaccard Measure (JM), Hausdorff Distance (HD) in mm.

|  | Lumen | | Media-adventitia | |
|---|---|---|---|---|
|  | JM | HD(mm) | JM | HD(mm) |
|  | 0.9335 | 0.0667 | **0.9509** | **0.0867** |
|  | 0.9381 | 0.0753 | 0.9479 | 0.1063 |
|  | 0.9351 | 0.6919 | 0.9497 | 0.0942 |
|  | 0.9275 | 0.8855 | 0.9473 | 0.1131 |
|  | 0.9371 | 0.6415 | 0.9359 | 0.1902 |
|  | 0.9381 | 0.0708 | 0.9449 | 0.1358 |
|  | **0.9412** | **0.0639** | 0.9387 | 0.1342 |
|  | 0.9357 | 0.0707 | 0.9496 | 0.1728 |
|  | 0.9331 | 0.0776 | 0.9376 | 0.1362 |
|  | 0.9357 | 0.0761 | 0.9401 | 0.1190 |
| **mean** | 0.9355 | 0.2720 | 0.9443 | 0.1289 |

The U-Net and IVUS-Net were used as the benchmark models. Table 2 compares JMs and HDs by different segmentation models for the test dataset. For both the lumen and the MA border, IVUS-U-Net++ achieved the best JM and the best HD compared with IVUS-Net and U-net.

Table 2 Comparison of different segmentation models with the test dataset. The evaluation measures are Jaccard Measure (JM) and Hausdorff Distance (HD). Data are shown as mean ± standard deviation.

|  | Lumen | | Media-adventitia | |
| --- | --- | --- | --- | --- |
|  | JM | HD(mm) | JM | HD(mm) |
| IVUS-U-Net++ | **0.9412**±0.0307 | **0.0639**±0.0436 | **0.9509**±0.0251 | **0.0867**±0.0622 |
| IVUS-Net | 0.9264±0.0362 | 1.0493±0.6130 | 0.9363±0.0385 | 1.6180±0.9875 |
| U-Net | 0.9304±0.0288 | 0.1078±0.1078 | 0.9403±0.0353 | 0.6278±1.2644 |

Figure 5 and Figure 6 illustrates segmentation results of three slices generated by U-Net (Group 1), IVUS-Net (Group 2) and IVUS-U-Net++ (Group 3), respectively.

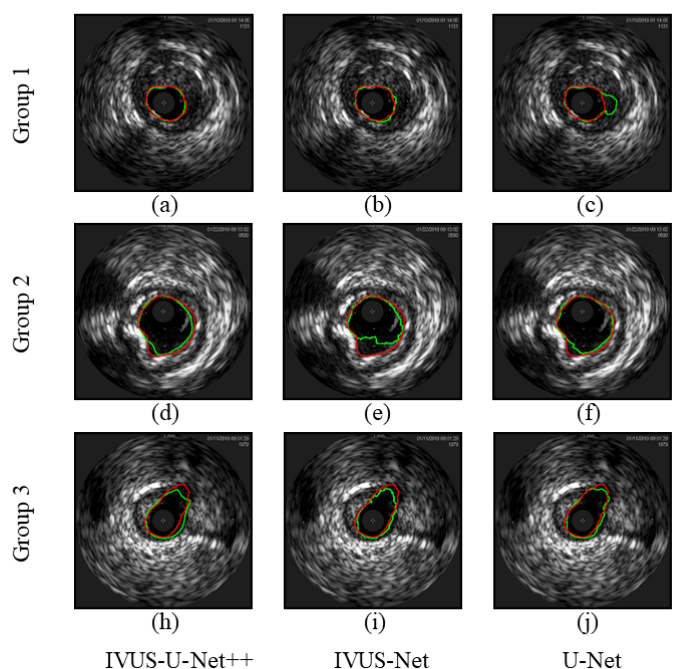

(a) (b) (c)
(d) (e) (f)
(h) (i) (j)
IVUS-U-Net++    IVUS-Net    U-Net

Figure 5 Segmentation results for the lumen border in the test dataset. The red contours represent the ground truth and the green contours represent the segmentation results from IVUS-U-Net++, IVUS-Net and U-Net. Figure c, e and h show the slices with the lowest JMs generated by U-Net, IVUS-Net and IVUS-U-Net++, respectively.

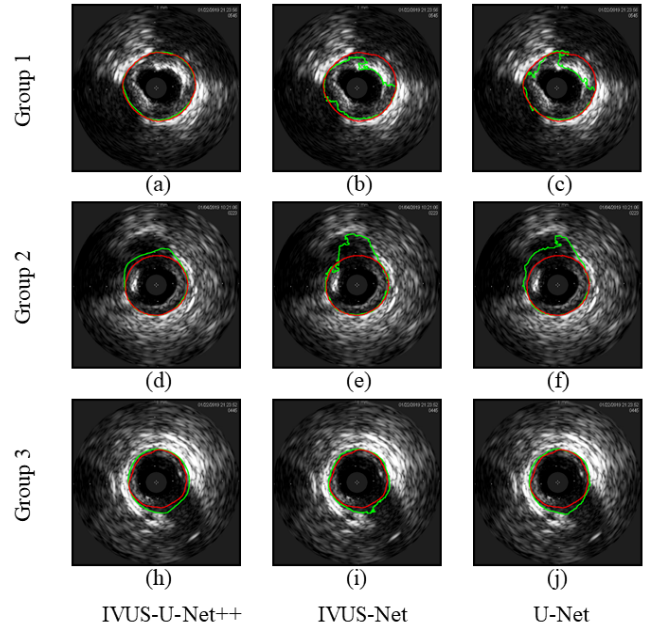

Figure 6 Segmentation results for the MA border in the test dataset. The red contours represent the ground truth and the green contours represent the segmentation results from IVUS-U-Net++, IVUS-Net and U-Net. Figure c, e and h show the slices with the lowest JMs generated by U-Net, IVUS-Net and IVUS-U-Net++, respectively.

The consistency and discrepancy between the commonly used clinical parameters measured from the ground truth and those from our IVUS-U-Net++ model was further evaluated to validate the accuracy of our model. Table 3 shows R (measured by the Pearson correlation analysis with IBM SPSS Statistics 20), mean absolute error, root mean square error and relative error. It can be observed that there is a high correlation with statistical significance for all the 12 clinical parameters. The scatter plots (Figure 7) and Bland-Altman analysis (Figure 8) further visualize the linear relationships and the consistencies for evaluating those clinical parameters. Figure 7 reflects the excellent linear correlations between them from the prediction and the ground truth. Figure 8 shows that with the 95% limits of agreement, most of the points are within the upper and lower consistency interval.

Table 3. Correlation and discrepancy between the clinical parameters measured from the ground truth and those from our IVUS-U-Net++ model. MAE, mean absolute error; RMSE, root mean square error; RE, relative error. For R measures, all Ps<0.01.

| Target | Metric | | | |
|---|---|---|---|---|
| | R | MAE | RMSE | RE(%)(min,max) |
| maximum EEM | 0.984 | 0.0806(mm) | 0.1131(mm) | (-0.0713, 0.1055) |
| minimum EEM | 0.991 | 0.0737(mm) | 0.0954(mm) | (-0.0597, 0.1107) |
| EEM CSA | 0.992 | 0.4000(mm$^2$) | 0.5473(mm$^2$) | (-0.1573, 0.1362) |
| maximum lumen | 0.986 | 0.0682(mm) | 0.1002(mm) | (-0.1166, 0.0670) |
| minimum lumen | 0.985 | 0.0779(mm) | 0.1009(mm) | (-0.0850,0.2660) |
| lumen CSA | 0.995 | 0.1845(mm$^2$) | 0.2482(mm$^2$) | (-0.1224,0.1965) |
| lumen eccentricity | 0.858 | 0.0347 | 0.0466 | (-0.6270,1.1489) |
| maximum plaque | 0.975 | 0.0805(mm) | 0.1074(mm) | (-0.1507,0.3494) |
| minimum plaque | 0.948 | 0.0514(mm) | 0.0698(mm) | (-0.7764,7.0623) |
| plaque plus | 0.981 | 0.4412(mm$^2$) | 0.5831(mm$^2$) | (-0.3145,0.1818) |
| plaque plus | 0.955 | 0.0420 | 0.058 | (-0.3880,0.3105) |
| plaque burden | 0.976 | 0.0163 | 0.0221 | (-0.1865,0.1108) |
| **mean** | 0.969 | 0.1292 | 0.1744 | - |

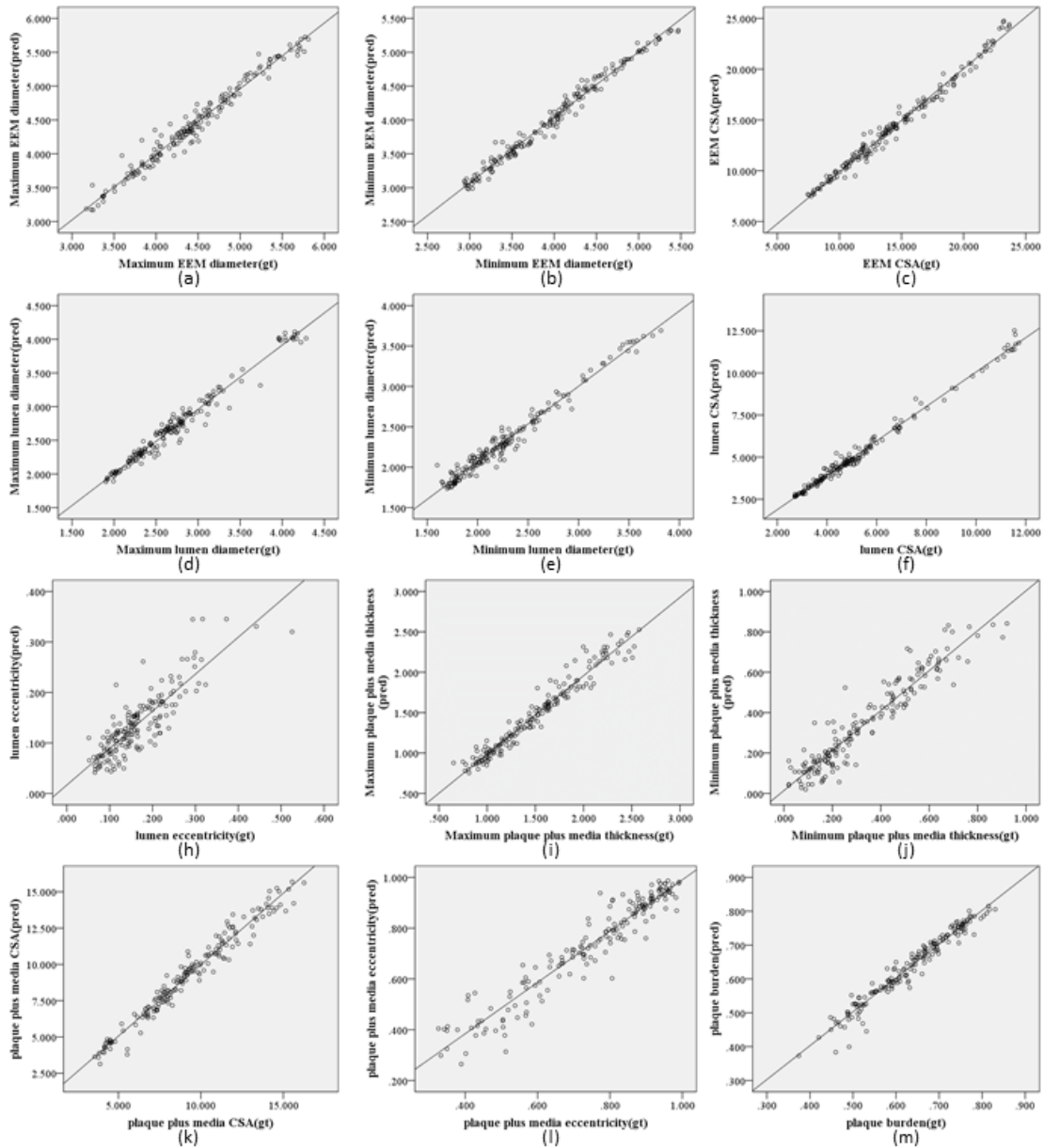

Figure 7. The scatter plots to visualize the linear correlations between clinical

parameters measured from the ground truth (horizontal axis) and those from the predicted results by IVUS-U-Net++ (vertical axis). (a)maximum EEM diameter, (b)minimum EEM diameter, (c)EEM CSA, (d)maximum lumen diameter, (e)minimum lumen diameter, (f)lumen CSA, (g)lumen eccentricity, (h)maximum plaque plus media thickness, (i)minimum plaque plus media thickness, (j) plaque plus media CSA, (k)plaque plus media eccentricity and (i)plaque burden.

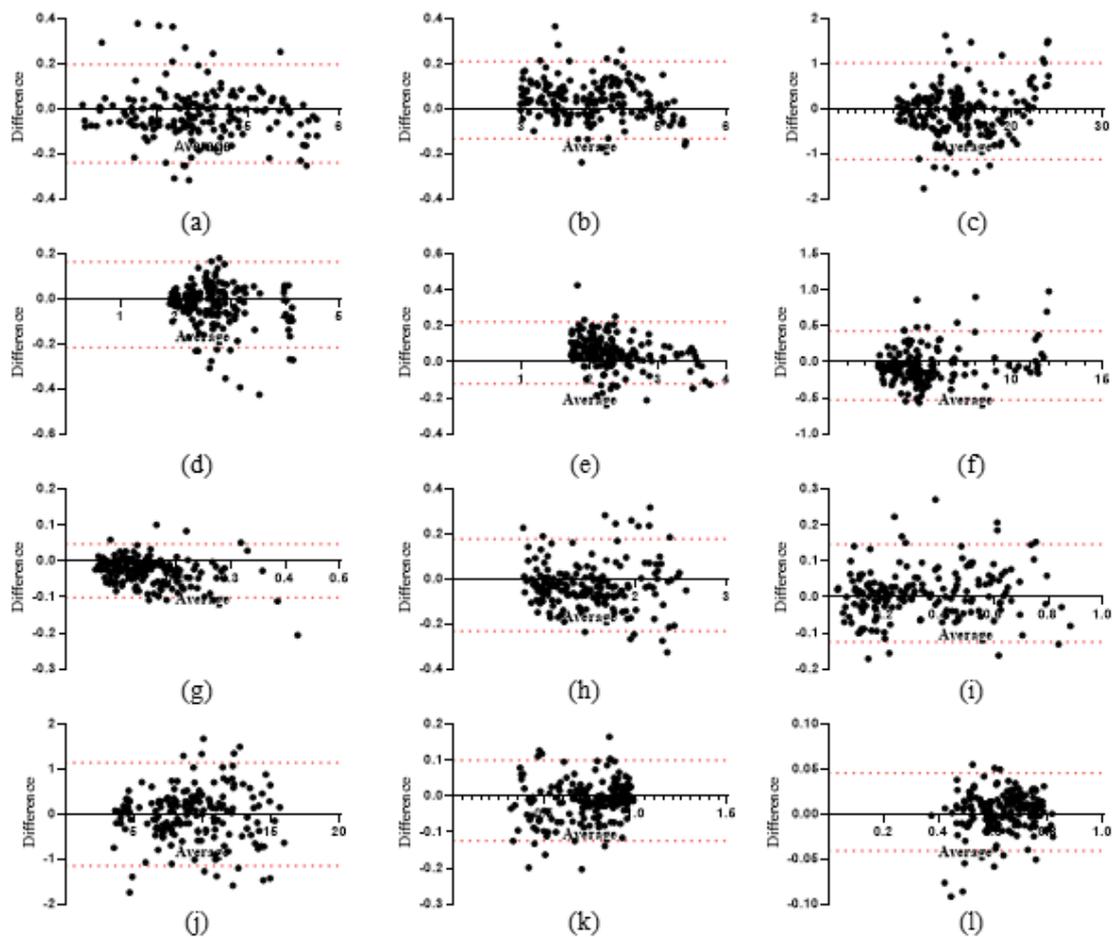

Figure 8. Bland-altman analysis (difference against average) between clinical parameters measured from the ground truth and those from the predicted results by IVUS-U-Net++. (a)maximum EEM diameter, (b)minimum EEM diameter, (c)EEM CSA, (d)maximum lumen diameter, (e)minimum lumen diameter, (f)lumen CSA, (g)lumen eccentricity, (h)maximum plaque plus media thickness, (i)minimum plaque plus media thickness, (j) plaque plus media CSA, (k)plaque plus media eccentricity and (i)plaque burden.

## 4. Discussion

A novel segmentation method has been proposed to extract the lumen and MA border in IVUS images. Pyramid feature extraction is added to the known U-Net++ model. The overall JMs and HDs were 0.9412, 0.0639 mm for the lumen border and 0.9509, 0.0867 mm for the MA border, respectively. There was an excellent agreement between the clinical parameters from our prediction results by IVUS-U-Net++ and the ground truth (all Ps<0.01). When compared with the other two state-of-the-art models, IVUS-U-Net++ achieved the best JM and the best HD, for both the lumen and MA border. In addition, our model could predict the segmentation result in 0.1203±0.0046 seconds for a single slice. It significantly reduces the annotation burden and has great promise for clinical use.

IVUS-U-Net++ shows significant advantages in retaining the global contours because of the multi-scale feature incorporation. It can be seen that In the Figure 5c, Figure 6e, and Figure 6f of Figure 6 there are significant differences in global contour shapes of lumen and MA between U-Net/ IVUS-Net and the ground truth. In contrast, IVUS-U-Net++ captures the global shapes with a consistent accuracy as shown in both Figure 5 and Figure 6. This is applicable even for the slices with the lowest JMs generated by IVUS-U-Net++ (Figure 5h and Figure 6h). The multiscale feature incorporation of the IVUS-U-Net++ significantly enlarges the receptive field with the techniques of pyramid features. Noteworthy, the IVUS-U-Net++ not only achieved the highest segmentation accuracy, but also produced smooth borders, similar to the radiologist's annotation.

The accuracy of our IVUS-U-Net++ model was further validated by the measurement of clinical parameters. There is a statistical significance ($P<0.01$) in all the 12 clinical parameters measured from the segmentation results from our model compared with the ground truth. A correlation of $R>0.90$ is obtained in 11 of these 12 clinical parameters. The Bland-Altman analysis shows that all of them maintain the 95% confidence interval. It should be noted that the lumen eccentricity calculated from the segmentation results doesn't agree well with that from the ground truth ($R=0.858$) because it relies on the precise calculations of both maximum lumen diameter and minimum lumen diameter, generating an accumulation of error.

## 5. Conclusion

A novel deep learning-based method has been proposed to extract the lumen and MA border from IVUS images. The quantitative evaluations in this feasibility study demonstrate that our method has a high accuracy and good robustness. It produces smooth contours and the computational efficiency is sufficient for its adoption in the clinical workflow. Our preliminary results are encouraging and warrant further investigations. The proposed method has the potential to be translated into the bedside.

**Conflict of interest**

The authors declare no conflicts of interest.

**Acknowledgment**

This research was supported by grants from Henan Science and Technology Development Plan 2020 (Project Number: 202102210384), 2019 Maker Space Incubation Project of Zhengzhou University of Light Industry (Project Number: 2019ZCKJ228), and Key Scientific Research Projects of Colleges and Universities in Henan Province (Project Number: 20A510014). This research was also supported in part by grants from the National Natural Science Foundation of China (Project Number: 81600279), from China Postdoctoral Science Foundation (Project Number: 2019M650075), and Pilot Project for Disruptive Technology (Biotechnology) from China National Center for Biotechnology Development (Project Number: 2020YFC1316700).